\documentclass[10pt,conference]{IEEEtran}

\usepackage[T1]{fontenc}      
\usepackage[utf8]{inputenc}
\usepackage{graphicx}
\usepackage{amsmath,amssymb}
\usepackage{booktabs}
\usepackage{multirow}
\usepackage{xcolor}
\usepackage{balance}          
\usepackage[hidelinks]{hyperref}
\usepackage{cite}
\usepackage{enumitem}
\usepackage{colortbl}
\usepackage{tikz}
\usetikzlibrary{positioning,arrows.meta,fit,backgrounds,calc}
\definecolor{drophl}{HTML}{F6E5E0}
\definecolor{dropln}{HTML}{A8412B}
\definecolor{teall}{HTML}{E2EDEB}

\definecolor{dA}{HTML}{EDE9E5}\definecolor{dB}{HTML}{E0C8BF}
\definecolor{dC}{HTML}{CE9C8B}\definecolor{dD}{HTML}{BC7259}
\definecolor{dE}{HTML}{A8412B}
\newcommand{\D}[1]{%
  \ifcase#1 \cellcolor{dA}\phantom{X}%
  \or \cellcolor{dB}\phantom{X}%
  \or \cellcolor{dC}\phantom{X}%
  \or \cellcolor{dD}\phantom{X}%
  \else \cellcolor{dE}\phantom{X}\fi}

\newcommand{\CorpusN}{121}
\newcommand{\BibN}{150}
\newcommand{\NLOne}{20}
\newcommand{\NLOneDirect}{13}
\newcommand{\NLOneTransfer}{2}
\newcommand{\NLOneContext}{5}
\newcommand{\NLTwo}{37}
\newcommand{\NLTwoDirect}{9}
\newcommand{\NLTwoTransfer}{21}
\newcommand{\NLTwoContext}{7}
\newcommand{\NLThree}{38}
\newcommand{\NLThreeDirect}{4}
\newcommand{\NLThreeTransfer}{21}
\newcommand{\NLThreeContext}{13}
\newcommand{\NLFour}{21}
\newcommand{\NLFourDirect}{4}
\newcommand{\NLFourTransfer}{12}
\newcommand{\NLFourContext}{5}
\newcommand{\NLFive}{5}
\newcommand{\NLFiveDirect}{0}
\newcommand{\NLFiveTransfer}{1}
\newcommand{\NLFiveContext}{4}
\newcommand{\NTotDirect}{30}
\newcommand{\NTotTransfer}{57}
\newcommand{\NTotContext}{34}
\newcommand{\NLThreeCertDirect}{1}
\newcommand{\NLThreeCertTransfer}{11}
\newcommand{\NLFourProofTransfer}{7}
\newcommand{\NLThreeRecent}{22}
\newcommand{\NRecent}{53}

\begin{document}

\bstctlcite{IEEEexample:BSTcontrol}

\title{SoK: Formal Methods for\\
       Fact-Checking and Information Integrity}

\author{
\IEEEauthorblockN{Nikolaos Kekatos\IEEEauthorrefmark{1},
Theodoros Nestoridis\IEEEauthorrefmark{1},
Charalampos Bratsas\IEEEauthorrefmark{3},
Charalampos Dimoulas\IEEEauthorrefmark{2},\\
Georgios Konstantinidis\IEEEauthorrefmark{4},
Georgios Malogiannis\IEEEauthorrefmark{1},
Michael Sirivianos\IEEEauthorrefmark{5},
Andreas Veglis\IEEEauthorrefmark{2}}
\IEEEauthorblockA{\IEEEauthorrefmark{1}Clone Systems, Cyprus \quad
\IEEEauthorrefmark{2}School of Journalism and Mass Communications, Aristotle University of Thessaloniki, Greece\\
\IEEEauthorrefmark{3}International Hellenic University, Greece \quad
\IEEEauthorrefmark{4}University of Southampton, UK \quad
\IEEEauthorrefmark{5}Cyprus University of Technology, Cyprus\\
Email: nkekatos@clone-systems.com, cbratsas@ihu.gr, babis@jour.auth.gr,\\
veglis@jour.auth.gr, michael.sirivianos@eecei.cut.ac.cy}
}

\maketitle

\begin{abstract}
An automated fact-checking system returns a label: the claim is true, or it
is false. In many such systems the verdict remains the primary output. What
is generally missing is a record of which document settled the question,
of what would have had to be different for the verdict to change, or of
whether the same claim, reworded, would have been judged the same way. We call the missing
piece a \emph{warrant}: a separate statement of what was guaranteed and on
what grounds. Formal methods produce evidence of this kind, and regulation
is beginning to ask for it, since the Digital Services Act and the AI Act
both call for auditable evidence about how systems behave.

Surveys of automated fact-checking are usually organised by pipeline stage,
and treat logic as one technique among many. We organise the field by
\emph{what is being formalised} instead, which gives five levels: the
claim, the reasoning, the system doing the checking, the ecosystem the
claim spreads through, and the regulatory obligation. Sorting \CorpusN{}
works into those levels, two patterns stand out. Most of the relevant
formal machinery already exists, but it was built for other domains and has
rarely been applied here, and the gap is widest for verifying the checking
system itself. Several stages of the routine professional fact-checkers
follow also have no stated correctness criterion, and two of them, writing a
claim in checkable form and correcting a verdict already published, are not
formally specified in any work we coded. We close with open problems,
each with a suggested first step.
\end{abstract}

\begin{IEEEkeywords}
fact-checking, formal methods, information integrity, disinformation,
neurosymbolic reasoning, runtime verification, certified robustness
\end{IEEEkeywords}

\section{Introduction}
\label{sec:intro}

Fact-checking is the practice of establishing whether a public claim is
true and publishing the finding together with the evidence for it. It is
carried out in two ways. Professional fact-checkers follow a documented
editorial routine of nine stages, running from monitoring public discourse
to correcting a verdict already published~\cite{graves2017anatomy,%
micallef2022trueorfalse,juneja2022human} (Section~\ref{sec:process}).
Automated fact-checkers implement a strict subset of that routine as a
pipeline~\cite{vlachos2014task,guo2022survey,zeng2021survey}: a claim is
detected, matched against claims checked before, supporting documents are
retrieved, and a model assigns a veracity label, sometimes with a generated
explanation.

The two differ in what reaches the reader. A published human check carries
its own justification: the reader sees which source settled the matter, on
what reading of the claim, and can disagree with the weighing. An automated
checker typically centres its output on a veracity label, sometimes
accompanied by retrieved evidence or a generated explanation. What is
generally missing is an independently checkable statement of which
retrieved document did the work, under what conditions the judgement would
flip, or whether the same claim in different words would have been judged
the same way. Because none of that
is recorded, nothing in the output distinguishes the cases where the system
is right from the cases where it is wrong. The confident mistakes look
exactly like the correct answers. What is missing is what we will call a
\emph{warrant}: an artefact, separate from the verdict and checkable on its
own, that says what was guaranteed and on what grounds.

The absence of such a warrant is particularly consequential in
fact-checking, for three reasons. The first is that someone is trying to
break the system. State-of-the-art synthetic-media detectors degrade catastrophically under
imperceptible perturbation, and fact-verification systems have been shown
to be manipulable by planted evidence~\cite{du2022synthetic}. The second is who reads the output. These verdicts
reach citizens making civic judgements, and an explanation that sounds
reasonable but does not follow from the evidence may appear justified
without being entailed by it. The third is that the law has
moved. The Digital
Services Act obliges very large online platforms to submit to independent
annual audits of their risk-mitigation measures~\cite{dsa2022}, and the AI
Act requires ex-ante conformity assessment for high-risk
systems~\cite{aiact2024}. Neither names fact-checking systems, and the DSA's audit duty falls on
very large platforms rather than on checkers. What both create is
\emph{demand for auditable evidence}, and formal methods are one way to
supply evidence stronger than a checklist: a checkable statement of what a
system guarantees, with evidence that the guarantee is discharged.

Formal methods (logic, automated reasoning, model checking, program
verification) exist precisely to produce such artefacts, and a substantial
body of work already applies them to fact-checking. We found no survey
that organises that body by what is being formalised. Five influential surveys structure the automated
fact-checking literature~\cite{guo2022survey,zeng2021survey,kotonya2020explainable,%
eldifrawi2024justification,dmonte2024claim}. All five organise it by
\emph{pipeline stage} (claim detection, evidence retrieval, veracity
prediction, justification) and treat logic-based methods as one technique
family competing with neural ones. None asks whether the system emitting
the verdict can itself be verified, and none engages with the
neural-network verification, runtime verification or probabilistic
model-checking literatures at all. Two adjacent systematizations organise by learning
paradigm~\cite{sok2023ml} and by human-centredness~\cite{sok2024chi}, and
inherit the same blind spot. A recent systematic survey does cover
symbolic methods in depth, including probabilistic logic and constraint
programming, but scopes itself to knowledge
graphs~\cite{rula2023kgfc}, which is one cell of the space we chart. Coding all eight against the
stages and levels used here leaves three rows empty for every one of them: the system, the ecosystem and the
regulatory obligation, together with the witness-, quantitative- and
observational-artefact families.

We organise the field differently: by \emph{what is being formalised}. This
single change of axis brings three previously disjoint literatures into
scope and yields a five-level assurance stack, from the formalisation of
claims themselves up to the formalisation of the regulatory obligations
that bind the systems processing them.

We ask three questions of that literature, and Section~\ref{sec:analysis}
answers each with a number from the coded corpus rather than an
impression.

\begin{description}[leftmargin=1.1em,itemsep=1pt,topsep=2pt]
\item[\textbf{RQ1}] \emph{What is being formalised?} Which objects of
information integrity have been subjected to formal or certifiable
assurance, and which artefact does each method produce?
\item[\textbf{RQ2}] \emph{Where is the adoption gap?} Which techniques are
instantiated directly on information-integrity systems, and which remain
transfer candidates from adjacent fields?
\item[\textbf{RQ3}] \emph{Where is the specification gap?} Which stages of
the professional routine lack a formally specified operation, and which
lack a correctness criterion for the operation they already have?
\end{description}

\emph{Contributions.}
\begin{enumerate}
\item \emph{A five-level assurance stack} (Section~\ref{sec:stack}) that
organises formal methods for information integrity by the object of
formalisation rather than by pipeline stage, together with a review of \CorpusN{}
works positioned within it (Section~\ref{sec:l1}--Section~\ref{sec:l5}). The two
lowest levels are mature; the third carries the largest pool of relevant
techniques but almost entirely as imported, uninstantiated tooling. The
top two are thinner, and the highest is almost entirely regulation and
context rather than method.
\item \emph{A grounded account of the professional fact-checking process}
(Section~\ref{sec:process}), against which we separate two kinds of gap:
constructing a checkable claim and correcting a published verdict are
absent as formally specified operations, while claim matching and
justification are implemented routinely but carry no stated correctness
criterion. Each has a plausible formal foundation, mature either in
restricted fact-checking settings or in adjacent formal-methods
research.
\item \emph{Eleven open problems} (Section~\ref{sec:open}), each stated with a
concrete first step.
\end{enumerate}

\section{Scope and Method}
\label{sec:scope}

\emph{Terminology.} We use \emph{warrant} as an umbrella term for
independently checkable assurance artefacts, not in the narrower Toulminian
sense of an inference-licensing rule.

\emph{What counts as a formal method.} We scope the term narrowly, to
\emph{techniques that produce a machine-checkable artefact with a defined
semantics}: a derivation, a proof object, a satisfying assignment, a
counterexample, a certificate, a monitor verdict. This excludes work that
is merely rigorous, statistically principled, or mathematically presented.
The test is whether the technique yields something a second, independent
tool could check. Under this definition a Horn-clause derivation qualifies
and an attention map does not. We also admit a small number of
\emph{adjacent} certifiable mechanisms, notably conformal prediction and
cryptographic provenance, which produce independently checkable artefacts
but fall outside a strict textbook definition of formal methods. Both are
flagged as boundary cases where they appear.

\emph{Sources and corpus.} We queried DBLP, the ACL Anthology, Semantic
Scholar and arXiv for the cross-product of formal-methods terms (model
checking, theorem proving, SAT, SMT, answer set programming, description
logic, argumentation, runtime verification, abstract interpretation, belief
revision) with information-integrity terms (fact-checking, claim
verification, misinformation, disinformation, influence operation,
information integrity), then followed citations forward and backward from
the resulting seed set. The window runs from Dung's foundational
argumentation paper~\cite{dung1995acceptability} to a cut-off of 31 August
2026. We did not run an exhaustive enumerate-then-screen search. From the
seed set we followed citations forward and backward to closure, iterating
until further queries stopped surfacing uncoded work. Per-stage screening
counts are not well defined for that strategy and we report none. We summarise the coding scheme here because the counts in
Section~\ref{sec:analysis} depend on it.

\emph{Corpus and bibliography.} \CorpusN{} works met the inclusion
criteria below and form the coded corpus. The bibliography is larger, at
\BibN{} entries, because the introduction, the process account
(Section~\ref{sec:process}) and the analysis (Section~\ref{sec:analysis}) cite
surveys, benchmarks and regulations that are not themselves coded. The two
numbers are generated separately and are not interchangeable.

\emph{What counts as a work.} A \emph{work} is a distinct bibliography
entry cited inside one of Section~\ref{sec:l1}--Section~\ref{sec:l5}. This denominator
deliberately includes standards, regulations and practitioner documents,
because part of the argument concerns what those artefacts do and do not
specify. Each entry carries one of three role codes. \emph{Direct}: a
formal or certifiable method applied to information integrity.
\emph{Transfer}: a mature method from another domain, mapped here to an
identified assurance need. \emph{Context}: a standard, regulation,
benchmark, survey, empirical study, or statement of professional
requirement. Where a work spans levels we assign it to the level of its
primary object and count it once. Role and artefact are tied together by
construction: a work is coded \emph{direct} or \emph{transfer} exactly
when it yields one of the artefacts of Section~\ref{sec:techniques}, so
Table~\ref{tab:matrix} accounts for every non-context work and the two
tables reconcile arithmetically rather than by inspection.
Every count in this paper is generated from that one file by script, so no
number here is maintained by hand. All eight authors coded the corpus
against a shared manual. A subset of works was coded independently by more
than one author, and disagreements were resolved by consensus before the
file was closed. Those duplicate codes reconciled assignments rather than
measuring agreement, so we report no reliability statistic.

\emph{Inclusion.} We include a work if it applies a formal method to some
stage of the information-integrity problem, \emph{or} if it supplies
tooling that the field demonstrably needs but has not adopted. The latter
category is essential, because Section~\ref{sec:l3} and Section~\ref{sec:l4} consist
largely of such work, and it is precisely what the \emph{transfer} code
isolates. We exclude purely neural methods, purely sociological studies,
and formal work on adjacent problems (spam, fraud, intrusion detection)
except where cited as methodological precedent.

\subsection{Threats to validity}
\label{sec:validity}

\emph{Non-enumerative search.} The corpus was assembled by seeding and
then snowballing to closure rather than by screening an enumerated result
set. That strategy has no reproducible per-stage yield, so corpus
completeness cannot be bounded from below and we do not claim it. Formal
work on information integrity is also published under terminology that does
not include ``fact-checking'', and cross-disciplinary vocabulary is the
most likely source of omission. The snowballing pass mitigates this without
eliminating it.

\emph{Boundary subjectivity.} \emph{Direct}, \emph{transfer} and
\emph{context}, and the assignment of a primary level to a work spanning
several, are judgements. They are recorded per work so a reader can
disagree with a specific row rather than with the aggregate.

\emph{Definition of a formal method.} We admit conformal prediction and
cryptographic provenance as boundary cases (Section~\ref{sec:scope}), which is
broader than a textbook definition. Both are flagged where they appear, so
the effect of excluding them is visible.

\emph{Negative claims.} Throughout, ``we identified no'' means absent
from this corpus at this cut-off. It is not a proof of non-existence, and
the additions made late in this work show the rate at which such claims
decay.

\emph{Process abstraction.} The nine stages synthesise published
workflows. No claim is made that every organisation implements all nine,
or implements them in that order.

\emph{Cut-off.} The corpus closes on 31 August 2026. Table~\ref{tab:era}
shows \NLThreeRecent{} of the \NLThree{} L3 works falling in the final
period, and \NRecent{} of \CorpusN{} across the corpus, so the findings
describe a fast-moving literature at one instant.

\section{The Fact-Checking Process}
\label{sec:process}

Verification presupposes a specification. You cannot check that a system is
correct without first saying what correct would mean. For fact-checking
there is no formal specification to appeal to, but there is something
close. Professional fact-checkers follow a routine; that routine has been
documented in detail by ethnographers, by interview studies, and by the
profession's own code of conduct, and it is the nearest thing the field has
to a statement of what the task requires. We set it out here, and then
measure the computational literature against it.

\subsection{The professional routine}

Graves' ethnography of US fact-checking organisations remains the canonical
account~\cite{graves2017anatomy}. Interview and workflow studies extend it
across countries and organisations~\cite{micallef2022trueorfalse,%
showmethework2025,juneja2022human}. The studies converge on nine stages: monitor, select, match, construct,
gather, adjudicate, justify, correct, intervene. For images and video the
gathering stage follows the profession's five pillars of verification,
namely provenance, source, date, location and
motivation~\cite{urbani2020pillars}. Fig.~\ref{fig:process}
sets them out and marks the four whose epistemic requirements are not
formalised.

\begin{figure*}[t]\centering
\footnotesize
\begin{tikzpicture}[
  fcstage/.style={draw=black!45, rounded corners=1.2pt, minimum height=9mm,
                  text width=15.5mm, align=center, inner sep=2pt,
                  font=\scriptsize, fill=black!4},
  fcdrop/.style ={fcstage, fill=drophl, draw=dropln, very thick},
  fcarr/.style  ={-{Latex[length=1.5mm]}, black!50},
  fccap/.style  ={font=\tiny\itshape, text=black!55, align=center,
                  text width=17mm}]
\def\dx{1.98}
\foreach \i/\code/\name/\sty in {%
  0/S1/Monitor/fcstage, 1/S2/Select/fcstage, 2/S3/Match/fcdrop,
  3/S4/Construct/fcdrop, 4/S5/Gather/fcstage, 5/S6/Adjudicate/fcstage,
  6/S7/Justify/fcdrop, 7/S8/Correct/fcdrop, 8/S9/Intervene/fcstage}{%
  \node[\sty] (n\i) at (\i*\dx,0) {\textbf{\code}\\[1pt]\name};}
\foreach \i in {0,...,7}{\pgfmathtruncatemacro{\j}{\i+1}
  \draw[fcarr] (n\i.east) -- (n\j.west);}
\node[fccap, below=1.0mm of n0] {ingest};
\node[fccap, below=1.0mm of n1] {check-worthiness};
\node[fccap, below=1.0mm of n2] {already checked?};
\node[fccap, below=1.0mm of n3] {scope the proposition};
\node[fccap, below=1.0mm of n4] {evidence, right of reply};
\node[fccap, below=1.0mm of n5] {house scale};
\node[fccap, below=1.0mm of n6] {replicable from sources};
\node[fccap, below=1.0mm of n7] {update on new evidence};
\node[fccap, below=1.0mm of n8] {debunk, prebunk};
\node[fcstage, minimum height=3.4mm, text width=3mm, inner sep=1pt]
  (lg1) at (1.1,-1.95) {};
\node[anchor=west, font=\scriptsize, text=black!65] at (1.35,-1.95)
  {addressed by the automated pipeline};
\node[fcdrop, minimum height=3.4mm, text width=3mm, inner sep=1pt]
  (lg2) at (8.6,-1.95) {};
\node[anchor=west, font=\scriptsize, text=dropln] at (8.85,-1.95)
  {epistemic requirement not formalised};
\end{tikzpicture}
\caption{The professional fact-checking process. Automation has
concentrated on ingest, triage and investigation. The four highlighted
stages carry the process's epistemic guarantees. Two of them, scoping a
claim and revising a published verdict, are largely absent from automated
pipelines. The other two, matching and justification, are implemented
routinely but with no stated correctness criterion (Section~\ref{sec:mismatch}).}
\label{fig:process}
\end{figure*}
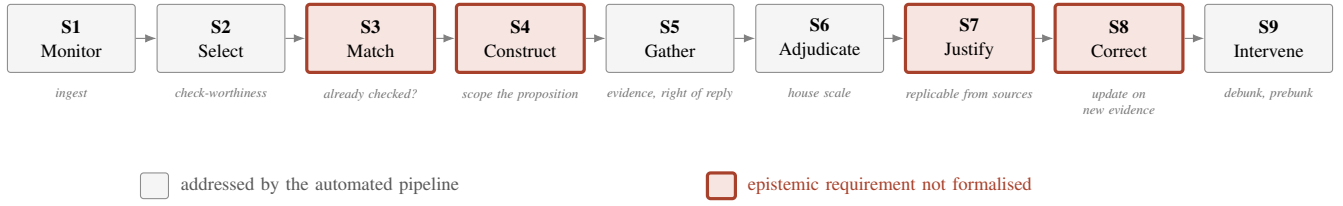
\begin{table}[t]\centering
\caption{Coverage of the main surveys of automated fact-checking and of two adjacent systematizations. Rows are the
process stages of Section~\ref{sec:process}, the assurance levels of
Section~\ref{sec:stack}, and the technique families of Section~\ref{sec:techniques}. Existing surveys concentrate on the middle of
the process and on the two lowest levels; none treats the system, the
ecosystem, or the regulatory obligation.}
\label{tab:surveys}
\scriptsize
\setlength{\tabcolsep}{3.2pt}
\renewcommand{\arraystretch}{1.06}
\newcommand{\rot}[1]{\rotatebox{90}{\scriptsize #1\hspace{1mm}}}
\newcommand{\yes}{$\bullet$}
\newcommand{\prt}{$\circ$}
\begin{tabular}{@{}l|ccccccccc@{}}
\toprule
& \rot{Guo et al.~\cite{guo2022survey}}
& \rot{Zeng et al.~\cite{zeng2021survey}}
& \rot{Kotonya \& Toni~\cite{kotonya2020explainable}}
& \rot{Eldifrawi et al.~\cite{eldifrawi2024justification}}
& \rot{Dmonte et al.~\cite{dmonte2024claim}}
& \rot{Corsi et al.~\cite{sok2023ml}}
& \rot{Razi et al.~\cite{sok2024chi}}
& \rot{Qudus et al.~\cite{rula2023kgfc}}
& \rot{\textbf{This paper}} \\
\midrule
\rowcolor{black!8}\multicolumn{10}{@{}l@{}}{\textbf{Process stage}}\\
S1--S2 Monitor, Select    & \yes & \yes &      &      & \yes & \yes & \yes & \prt & \yes \\
S3 Match                  & \yes & \prt &      &      & \prt &      &      &      & \yes \\
S4 Construct              &      &      &      &      &      &      &      &      & \yes \\
S5--S6 Gather, Adjudicate & \yes & \yes & \prt & \prt & \yes & \yes & \prt & \yes & \yes \\
S7 Justify                & \yes & \prt & \yes & \yes & \prt &      &      & \prt & \yes \\
S8 Correct                &      &      &      &      &      &      &      &      & \yes \\
S9 Intervene              & \prt &      &      &      &      &      &      &      & \yes \\
\midrule
\rowcolor{black!8}\multicolumn{10}{@{}l@{}}{\textbf{Assurance level}}\\
L1 Object                 & \yes & \yes & \prt &      & \yes & \prt &      & \yes & \yes \\
L2 Reasoning              & \prt &      & \yes & \yes & \prt &      &      & \yes & \yes \\
L3 System                 &      &      &      &      &      & \prt & \prt &      & \yes \\
L4 Ecosystem              &      &      &      &      &      &      &      &      & \yes \\
L5 Governance             &      &      &      &      &      &      &      &      & \yes \\
\midrule
\rowcolor{black!8}\multicolumn{10}{@{}l@{}}{\textbf{Technique family}}\\
Proof-producing           & \prt &      & \yes & \prt & \prt &      &      & \yes & \yes \\
Witness-producing         &      &      &      &      &      &      &      &      & \yes \\
Certificate-producing     &      &      &      &      &      & \prt &      &      & \yes \\
Quantitative              &      &      &      &      &      &      &      &      & \yes \\
Observational             &      &      &      &      &      &      &      &      & \yes \\
\bottomrule
\end{tabular}
\par\vspace{3pt}
{\scriptsize\raggedright
\yes\ treated as a topic in its own right, \prt\ mentioned but not
systematised, blank absent.\par}
\end{table}

Two features of this routine deserve emphasis because they do not survive
translation into the computational literature. First, the IFCN Code of
Principles (the field's own standard, binding on certified
signatories) consists of five commitments of which four constrain the
\emph{process} rather than the output: nonpartisanship and fairness,
transparency of sources, transparency of methodology, and an open
corrections policy~\cite{ifcn_code}. A code of conduct that regulates
procedure is, structurally, a specification. Second, commitment three
requires that sources be given ``in enough detail that readers can
replicate'' the check. That is a requirement of \emph{auditability}, stated by practitioners
about their own output. It is not yet a soundness requirement: entailment
of the justification by the evidence, which we develop in
Section~\ref{sec:l3}, is our proposed formal strengthening of it, not a
restatement.

Third, verdicts are assigned on \emph{house} scales that are not
commensurable: a six-point Truth-O-Meter, a five-point scale ending in
\emph{uncheckable}~\cite{eufactcheck}, narrative verdicts, and many local
variants, with no agreed mapping between them (Section~\ref{sec:l3}).

\subsection{The automated pipeline}

The computational literature formalises a strict subset. Following the
original task formulation~\cite{vlachos2014task} and the consensus of the
surveys~\cite{guo2022survey,zeng2021survey}, an automated fact-checking
system comprises: \emph{claim detection}, usually decomposed into claim
spotting and check-worthiness ranking~\cite{hassan2017claimbuster,%
nakov2022checkthat,sundriyal2023claimdetection}; \emph{claim matching}
against previously checked claims, a stage added by operational practice
rather than by theory~\cite{shaar2020knownlie}; \emph{evidence retrieval};
\emph{veracity prediction}; and \emph{justification
production}~\cite{kotonya2020explainable,eldifrawi2024justification}. The
framing throughout is assistive rather than
autonomous~\cite{nakov2021assisting}. A representative modular realisation is
OpenFactCheck~\cite{iqbal2025openfactcheck}, which consolidates competing
systems into a configurable chain of \texttt{claim\_processor},
\texttt{retriever} and \texttt{verifier}. We use it below as a concrete
reference point, not as a definition of the task.

\subsection{The mismatch}
\label{sec:mismatch}

Overlaying the two accounts produces our second finding, which is a claim
about \emph{specification} rather than about attention: \emph{the stages
carrying the process's epistemic guarantees are either absent from the
computational pipeline or present in it with no stated correctness
criterion}. The distinction matters. Claim matching and justification
production are established automated tasks with substantial literatures. What neither has is a criterion under which its output is \emph{correct}.
Claim construction and correction differ: they are largely missing as
specified operations at all. Table~\ref{tab:omitted} sets out all four
with the criterion each lacks and the formalism that supplies it.

\begin{table}[t]\centering
\caption{The four stages whose epistemic requirements are not formalised.
S4 and S8 are largely absent from automated pipelines; S3 and S7 are
implemented routinely but with no stated correctness criterion.}
\label{tab:omitted}
\scriptsize
\setlength{\tabcolsep}{3pt}
\begin{tabular}{@{}lp{1.45cm}p{1.9cm}p{2.3cm}@{}}
\toprule
\textbf{Stage} & \textbf{In automation} & \textbf{Missing criterion} & \textbf{Candidate formalism} \\
\midrule
S3 Match & established & semantic claim equivalence & natural logic lifted to a matching relation \\
S4 Construct & largely absent & faithful scoped proposition & autoformalisation; temporal and numeric calculi \\
S7 Justify & established & evidence entails the justification & proof-producing reasoning \\
S8 Correct & largely absent & principled revision under new evidence & belief revision \\
\bottomrule
\end{tabular}
\end{table}

Two of these need a qualification, because the gap is one of
\emph{soundness criteria} rather than of attention. OpenFactCheck's claim
processor includes a \emph{decontextualise} step and its verifier an
\emph{edit} step~\cite{iqbal2025openfactcheck}: recognisably partial
attacks on S4 and S8. Decontextualisation resolves references so a
claim can stand alone. It does not fix the quantifier, interval,
population, measure or data revision that make a claim checkable, and
\emph{edit} revises a generated text towards the evidence. It is not
revision of a \emph{published verdict} when the evidence later changes. The
field has recognised both needs and addressed them heuristically. What is
missing is a criterion for when either operation is correct.

The starkest case is S8. Correction can be modelled as belief revision:
how a rational agent updates a theory when new information contradicts it
is the AGM problem~\cite{alchourron1985agm}. Newsroom correction is not
reducible to it, since it also involves versioning, provenance and
editorial rule, with
four decades of subsequent development in dynamic
doxastic logic~\cite{vanbenthem2007dynamic}. Fact-checking organisations
perform this operation by hand, dozens of times a year, under a published
policy. Two adjacent literatures exist and neither closes the gap. Cognitive science studies how \emph{readers} update on a
correction~\cite{swire2021timing}, a different object from the checker's
own corpus, and the AGM tradition has developed credibility-limited
revision, which relaxes the success postulate precisely because not all
incoming information deserves acceptance~\cite{credibility_limited2023}.
That is the formalism S8 calls for, and we identified no work in our corpus applying
it to the maintenance of a published fact-check corpus.

Finally, a point about method. Graves finds that checkers establish truth by
reconciling a claim against a web of accepted sources rather than by direct
comparison with the world~\cite{graves2017anatomy}. That is why the
consistency-based formalisms surveyed in Sections~\ref{sec:l1}
and~\ref{sec:l2}, from knowledge-graph paths through description-logic ABox
checking to satisfiability encodings of incident reports, fit this domain.
They match what practitioners already do.

\section{The Assurance Stack}
\label{sec:stack}

We organise the literature by the object of formalisation, which gives five
levels.

\emph{L1: The object.} What kind of thing is a claim, formally, and when
does a piece of evidence bear on it?

\emph{L2: The reasoning.} Is the step from evidence to verdict a
derivation, and can an independent checker replay it?

\emph{L3: The system.} Is the artefact performing the check itself
verified (robust under specified perturbation, monitored at runtime, sound
in its explanations, and trustworthy in its provenance handling)?

\emph{L4: The ecosystem.} Can we reason formally about the environment
in which claims propagate and interventions act?

\emph{L5: Governance.} Can the obligations imposed by regulation be
expressed as specifications and discharged with evidence?

The levels are objects of increasing system scope, not a dependency chain. A diffusion model can be built without a verified checker, and a regulatory
obligation can be formalised without a claim calculus. Five is not an
arbitrary number: the levels enumerate the objects whose properties a
verification technique can be about in this domain, namely an individual
information object, an inference over such objects, the computational
system performing that inference, the multi-agent environment the claim
moves through, and an obligation imposed from outside it. Every work in
the corpus was assignable to one of the five as its primary object, and we
encountered none that required a sixth. A single paper may
span several levels, and we tag by primary object. As Section~\ref{sec:analysis}
shows, effort is not distributed accordingly. The upper levels are served
by mature tooling that this field has not taken up.

\subsection{Which techniques apply, and what each yields}
\label{sec:techniques}

Formal methods are not interchangeable, and the useful axis of comparison
is not power but \emph{output}. An information-integrity system needs
something it can hand to an auditor, a journalist or a regulator.
Table~\ref{tab:guarantees} states the five artefact classes with the form
of guarantee each yields, the assumption it rests on, and the way it fails;
Table~\ref{tab:techniques} then maps the technique families onto them with
tools and costs, and Table~\ref{tab:examples} gives one concrete
instantiation per family, so that the artefact each produces can be judged
against a specific question rather than a category name. The failure column carries most of the weight, because
every row can be discharged and still be useless, for a different reason
per row.

\begin{table*}[t]\centering
\caption{The guarantee taxonomy. The axis that matters for information
integrity is not power but what the method hands to a second party. Each
row states the artefact, the form of the guarantee, what must hold for it
to mean anything, and how it fails in practice. The last column is the one
this literature most often leaves implicit, since every row can be
satisfied and still be useless.}
\label{tab:guarantees}
\footnotesize
\setlength{\tabcolsep}{5pt}
\renewcommand{\arraystretch}{1.15}
\begin{tabular}{@{}p{30mm}p{31mm}p{50mm}p{52mm}@{}}
\toprule
\textbf{Assurance artefact} & \textbf{Guarantee form} &
\textbf{What must hold} & \textbf{How it fails} \\
\midrule
Proof, derivation
& $E \vdash C$
& The formalisation is faithful to the claim a human wrote.
& Silent mistranslation. The derivation is valid and answers a different
  question, so the output is a \emph{confidently wrong certificate}
  (Section~\ref{sec:l2}). \\
Witness, counterexample
& $\exists x.\,\neg P(x)$
& The model is complete enough that a witness in the encoding is a witness
  in the world.
& Model mismatch. The minimal unsatisfiable subset localises a conflict in
  the encoding rather than in the evidence. \\
Certificate
& $\forall x \in S.\,P(x)$
& $S$ is the perturbation set an adversary actually uses.
& Wrong specification set, not unsoundness: an $L_\infty$ ball may cover
  only a limited subset of the manipulations that matter here
  (Section~\ref{sec:l3}). \\
Quantitative bound
& $\Pr[\varphi] \bowtie p$
& The model is calibrated against measured behaviour, and tractable at
  realistic scale.
& Uncalibrated, or intractable and silently replaced by simulation
  (Section~\ref{sec:l4}). \\
Runtime verdict
& $\tau \models \varphi$
& The events the property mentions are observable in the deployed system.
& Incomplete instrumentation. The monitor is sound over the trace it sees
  and the trace omits the failure. \\
\bottomrule
\end{tabular}
\par\raggedright\vspace{2pt}
{\footnotesize Conformal prediction sits in the quantitative row as a
boundary case (Section~\ref{sec:scope}): its guarantee is
$\Pr[L \le \epsilon] \ge 1-\alpha$, its assumption is exchangeability
rather than calibration, and its characteristic failure is distribution
shift.}
\end{table*}

\begin{table*}[t]\centering
\caption{Formal verification techniques applicable to information
integrity, grouped by the artefact each produces. \textsc{l} refers to the
assurance stack of Section~\ref{sec:stack}, and the artefact classes are those of
Table~\ref{tab:guarantees}. Maturity is assessed for this domain
specifically, not in general: several mature techniques have never been
instantiated here.}
\label{tab:techniques}
\scriptsize
\setlength{\tabcolsep}{4pt}
\begin{tabular}{@{}p{26mm}p{27mm}p{28mm}p{45mm}p{25mm}c@{}}
\toprule
\textbf{Technique family} & \textbf{Artefact produced} & \textbf{Cost / limitation} & \textbf{Application in this domain} & \textbf{Tools} & \textsc{l} \\
\midrule
\multicolumn{6}{@{}l@{}}{\emph{Proof-producing: the answer is a replayable derivation}}\\
Answer set programming      & stable model + derivation   & grounding blow-up           & verdict from rules over KG evidence; explanation falls out of the proof & clingo, DLV & 2 \\
Description logic           & entailment or clash         & decidable fragments only    & claim consistency against a reference ontology & HermiT, ELK & 1 \\
Horn-rule rewriting         & rule chain                  & needs a rule base           & human-readable evidence for a candidate fact & ExFaKT, AMIE & 2 \\
Interactive proof           & machine-checked proof       & very high human cost        & metatheory of the pipeline; rarely worth it here & Coq, Isabelle & 2 \\
\addlinespace[2pt]
\multicolumn{6}{@{}l@{}}{\emph{Witness-producing: the answer localises the conflict}}\\
SAT / MaxSAT                & model, or \textsc{mus}      & NP-hard, practical at scale & incident-report consistency, completion, fusion & CaDiCaL, RC2 & 1--2 \\
SMT                         & model or proof              & theory-dependent            & numeric and temporal claim scope; autoformalisation target & Z3, cvc5 & 1 \\
Symbolic model checking     & counterexample trace        & state explosion             & moderation policy as a transition system & nuXmv, SPIN & 4 \\
Strategic logic (ATL)       & strategy or counterexample  & doubly exponential          & can $k$ inauthentic accounts force an outcome? & MCMAS & 4 \\
\addlinespace[2pt]
\multicolumn{6}{@{}l@{}}{\emph{Certificate-producing: one-sided guarantee over a perturbation set}}\\
Complete NN verification    & certificate or counterexample & exponential; small nets   & detector robustness; adversarial example synthesis & Marabou & 3 \\
Bound propagation           & one-sided robustness bound  & incomplete (``unknown'')    & scalable certified robustness for deployed detectors & $\alpha,\!\beta$-CROWN, ERAN & 3 \\
\addlinespace[2pt]
\multicolumn{6}{@{}l@{}}{\emph{Quantitative: the answer is a probability, not a yes or no}}\\
Probabilistic model checking& probability bound (PCTL)    & intractable at real scale   & \emph{verified interventions}: cascade probability under prebunking & PRISM, Storm & 4 \\
Statistical model checking  & probability $\pm$ confidence& weaker guarantee            & the same properties at realistic network size & PRISM, Plasma & 4 \\
\addlinespace[2pt]
\multicolumn{6}{@{}l@{}}{\emph{Observational: the answer is per execution, not over all executions}}\\
Runtime verification        & monitor verdict per trace   & no static guarantee         & crisis-mode SLAs, bounded FP rate, drift alarms & MonPoly, TeSSLa & 3 \\
Conformal prediction$^{\dag}$& coverage guarantee         & exchangeability only        & calibrated abstention; factuality control & n/a & 3 \\
\bottomrule
\end{tabular}
\par\raggedright\vspace{2pt}
{\scriptsize $^{\dag}$Statistical rather than formal; included because it
occupies the complementary regime discussed in Section~\ref{sec:l3}, supplying
probabilistic coverage or risk guarantees under distributional
assumptions.}
\end{table*}

\begin{table*}[t]\centering
\caption{One indicative instantiation per technique family. Each row names
a concrete question in this domain, the object the method is actually given,
and the artefact a second party receives. The point of the last column is
that the artefact is checkable without rerunning the system: a derivation can be
replayed, a witness can be inspected, a certificate can be audited against
its perturbation set.}
\label{tab:examples}
\scriptsize
\setlength{\tabcolsep}{4pt}
\begin{tabular}{@{}p{26mm}p{58mm}p{66mm}@{}}
\toprule
\textbf{Technique family} & \textbf{Indicative instance} & \textbf{What the second party receives} \\
\midrule
\multicolumn{3}{@{}l@{}}{\emph{Proof-producing}}\\
Answer set programming & ``$X$ held office in $Y$ during 2019'': office-holding intervals and succession rules over the reference graph & the stable model containing the office-holding atom, plus the rule chain that derived it; the explanation is the proof, not a post-hoc rationalisation \\
Description logic & the claim types a municipality as the subject of a property whose domain is a person & a clash, with the minimal set of ontology axioms and asserted triples that cannot hold together \\
Horn-rule rewriting & a candidate fact with no direct triple in the graph & a rule chain grounded in retrievable sentences, so the reader sees which mined regularity carried the inference \\
Interactive proof & the operator that aggregates per-source evidence into a verdict & a machine-checked proof of a metatheorem, e.g.\ that adding a corroborating source can never lower a verdict \\
\addlinespace[2pt]
\multicolumn{3}{@{}l@{}}{\emph{Witness-producing}}\\
SAT / MaxSAT & five incident reports giving mutually inconsistent casualty counts and timings & a minimal unsatisfiable subset naming the two reports that cannot both hold, rather than a global ``inconsistent'' \\
SMT & ``inflation fell by more than three points between Q1 and Q3'' against the published series & a model of the scope under which the claim is true, or a proof that no such scope exists \\
Symbolic model checking & a platform's moderation policy as a transition system & a counterexample trace: an item demoted and then restored without the review step the policy requires \\
Strategic logic (ATL) & $k$ coordinated inauthentic accounts against a ranking mechanism & a strategy witnessing that the coalition can force the outcome whatever honest users do, or a proof that it cannot \\
\addlinespace[2pt]
\multicolumn{3}{@{}l@{}}{\emph{Certificate-producing}}\\
Complete NN verification & a detector and a declared edit budget around one claim & a certificate that no edit within the budget flips the label, or the specific paraphrase that does \\
Bound propagation & the same detector at deployment scale & a certified radius per input, or an explicit ``unknown'' (incomplete, never unsound) \\
\addlinespace[2pt]
\multicolumn{3}{@{}l@{}}{\emph{Quantitative}}\\
Probabilistic model checking & a diffusion model with and without a prebunking intervention & a bound on the probability that a cascade exceeds $n$ reshares within $t$ steps, for both arms: a verified comparison rather than a simulated one \\
Statistical model checking & the same property on a network too large to solve exactly & the probability with a confidence interval, from sampled executions \\
\addlinespace[2pt]
\multicolumn{3}{@{}l@{}}{\emph{Observational}}\\
Runtime verification & a crisis-mode service level: every flagged item reviewed by a human within 30 minutes & a verdict per execution, with the prefix of the trace that violates the property \\
Conformal prediction$^{\dag}$ & a deployed checker that may abstain & a threshold such that the error rate among accepted items stays below $\epsilon$ with probability $1-\alpha$, under exchangeability \\
\bottomrule
\end{tabular}
\par\raggedright\vspace{2pt}
{\scriptsize $^{\dag}$Statistical rather than formal, as in
Table~\ref{tab:techniques}.}
\end{table*}

Proof-producing methods return a derivation and so fit L2, where the
requirement is that a verdict be replayable. Witness-producing methods
return an assignment or a counterexample, and their value here is
\emph{localisation}. A minimal unsatisfiable subset names the assertions
responsible rather than reporting global failure. Certificate-producing
methods return a one-sided guarantee over a declared perturbation set and
are the techniques that directly support universal
robustness claims over a declared perturbation set. Their
limitation is the specification, not soundness (Section~\ref{sec:l3}).
Quantitative methods return a rate, a bound or a risk rather than a yes or
no, which suits L4 questions about how likely a cascade is. L4 also hosts qualitative formalisms such as epistemic logic and ATL. Observational methods
return a verdict per execution rather than over all executions, trading
universality for the ability to run on systems that cannot be analysed
statically, which in practice means any pipeline containing a language
model.

\section{L1: Formalising Claims and Evidence}
\label{sec:l1}

\subsection{Knowledge-graph triple verification}

The simplest formalisation of a claim is a triple, and the simplest question
to ask of one is whether a reference graph already supports it.
The founding line treats a claim as a triple and truth as a structural
property of a reference graph: shortest-path
specificity~\cite{ciampaglia2015knowledgelinker}, discriminative predicate
paths~\cite{shi2016predpath}, maximum flow~\cite{shiralkar2017stream},
corroborative multi-source search~\cite{syed2019copaal} and text--graph
hybrids~\cite{hybridfc2024}. A dedicated survey covers the subfield in
depth~\cite{rula2023kgfc} and we defer to it.

This line established that veracity can be a computed property rather than
a learned one. It also exposed the field's most persistent semantic
problem. Under the open-world assumption, the absence of a supporting path
is not evidence of falsity, yet path-based scores are routinely interpreted
as though it were. We return to this in Section~\ref{sec:open}.

\subsection{Description logics and ontological consistency}

A claim can also be wrong by contradicting what is already known rather
than by lacking support, which is a consistency question and not a retrieval
one. A second formulation asks whether a candidate assertion is \emph{consistent}
with the TBox and ABox of a reference ontology, reducing fact-checking to a
decidable reasoning problem~\cite{swj_logicalconsistency}. Because real
knowledge graphs are inconsistent, this requires machinery for reasoning
under contradiction: paraconsistent semantics for description
logics~\cite{ma2007paraconsistent} and, more recently, a systematic survey
of inconsistency handling in knowledge-graph
reasoning~\cite{kginconsistency2025}. On the engineering side, SHACL
supplies a declarative constraint layer, and learned approximations to
consistency checking now claim order-of-magnitude speedups over classical
reasoners at 95\% agreement~\cite{glamor2025}: a development that must be
treated carefully, since an approximate consistency check is not a
certificate.

\subsection{Natural logic}

Translating a sentence into logic is where faithfulness is usually lost, so
one line avoids the translation and reasons over the sentence itself.
Natural logic reasons over surface forms via monotonicity and a small
algebra of entailment relations~\cite{maccartney2009natlog}, avoiding
translation into a formal language entirely. ProoFVer determines veracity \emph{solely} as a
function of a sequence of operator-labelled lexical
mutations~\cite{krishna2022proofver}. The consequence is important out of
proportion to the accuracy gain. The explanation is faithful \emph{by
construction}, because the verdict is a deterministic function of the
displayed proof. ProoFVer also improves by 13.2 points over the next best
model on counterfactual instances, suggesting that the constraint buys
robustness as well as transparency. The line continues with
question-answering formulations~\cite{aly2023qanatver}, tabular
extensions~\cite{aly2025tabver}, and zero-shot combinations with
LLMs~\cite{zeroshot_natlog2024}.

\subsection{Executable logical forms}

Where a claim quantifies, aggregates or compares, semantic similarity is
insufficient and symbolic execution is required. TabFact established the
benchmark~\cite{chen2020tabfact} and LogicalFactChecker derives an
executable program by semantic parsing~\cite{zhong2020logicalfactchecker}.
For numerical claims over real-world corpora~\cite{quantemp2024} the
difficulty is less the arithmetic than recovering the intended measure. The
most explicit attack on the representation itself extends first-order logic
over knowledge graphs with comparison predicates and counting quantifiers,
pairs 43{,}821 claims with formulas, and supplies an executable
prover~\cite{hao2026folxkg}: evidence that quantification, comparison and
counting can be made formally explicit for graph-grounded claims.

\subsection{Temporal and numeric claim calculi}

Triples are timeless; claims are not. Temporal fact-checking assigns
validity intervals to assertions~\cite{temporalfc2023}, but no accepted
formalism exists for the temporal scope of a natural-language claim: the
interval over which ``unemployment is falling'' is asserted, and the data
revision against which it is to be judged. This is the technical core of
the missing S4 stage identified in Section~\ref{sec:mismatch}.

\section{L2: Formalising the Reasoning}
\label{sec:l2}

\subsection{Answer set programming}

ASP offers non-monotonic, default and aggregate reasoning with a mature
solver ecosystem. Ahmadi et al.\ cast fact-checking as inference in a
probabilistic answer-set program, obtaining interpretable explanations as a
by-product of the derivation~\cite{ahmadi2019paspfc,plingo2022}. Related
work covers review authenticity~\cite{asp_reviews2021} and LLM-to-ASP
coupling for text reasoning~\cite{ishay2023coupling}.

\subsection{Rule mining and Horn-clause explanation}

ExFaKT rewrites a hard-to-verify fact into a set of easier-to-spot facts
using Horn rules drawn from background knowledge. The rewriting \emph{is}
the human-readable explanation~\cite{gadelrab2019exfakt}. The rules
themselves can be mined: AMIE and its successors mine Horn rules under
incomplete evidence~\cite{galarraga2020amie3}, while RuDiK mines both
positive and negative rules and is explicitly robust to errors in the
underlying graph~\cite{ortona2018rudik}. The negative rules matter,
because refutation and confirmation are not symmetric tasks. Recent work
brings neural methods to rule mining without abandoning the symbolic
output~\cite{neurosymbolic_rulemining2024}.

\subsection{Computational argumentation}

Argumentation's native output is a debate structure rather than a label. Dung's abstract frameworks define
acceptability semantics over an attack graph~\cite{dung1995acceptability};
structured extensions such as ASPIC$^+$~\cite{modgil2014aspic} and
bipolar or value-based frameworks add internal structure and preference.
Applications to veracity include argumentation-based explainable
fact-checking~\cite{kotonya2020argumentation}; logic-programme induction over
multimodal evidence~\cite{logicdm2023} and, notably, misinformation
identification via \emph{critical question
answering}~\cite{criticalquestions2025}, which operationalises argumentation
schemes as an interrogation of a claim's warrant. This last is directly
usable as the formal backbone of a media-literacy intervention, a rare
property in this literature. Contrastive explanation over argumentation
conclusions has also been formalised~\cite{contrastive2021}.

\subsection{Inconsistency, paraconsistency and belief revision}

Real evidence sets contradict themselves. Inconsistency measurement treats
this as a quantity rather than a failure
state~\cite{hunter2006measuring,thimm2013inconsistency}; decomposition into
minimal unsatisfiable subsets converts a contradiction into a localised,
explainable conflict~\cite{marquessilva2014mus}; and paraconsistent
inference relations permit useful conclusions to be drawn from an
inconsistent base~\cite{paraconsistent2022}. The applications already
enumerated in this literature include news reports and integrity
constraints, but the connection has not been made in the other direction.
Belief revision~\cite{alchourron1985agm} remains, as noted, entirely absent
from the fact-checking literature despite being the formal theory of its
S8 stage.

\subsection{Probabilistic logic and truth discovery}

Evidence for a claim arrives from several sources of unequal reliability and
rarely agrees with itself, so the encoding has to carry weights rather than
hard constraints. Markov logic networks combine first-order structure with weighted
uncertainty~\cite{richardson2006mln}; probabilistic soft logic relaxes
truth to $[0,1]$ under {\L}ukasiewicz semantics, making MAP inference a
convex optimisation~\cite{bach2017psl}. Knowledge Graph Identification
performs error correction, deduplication, link prediction and inconsistency
detection jointly in PSL~\cite{pujara2013kgi}. Orthogonally, the truth-discovery
literature jointly infers source reliability and claim
veracity~\cite{dong2009truthdiscovery,li2016truthsurvey}. It supplies the
weights a soft-constraint encoding of multi-source evidence needs, and the
formal-methods work has not taken it up.

\subsection{Neurosymbolic methods}

This is where most work since 2023 has been published. ProgramFC
decomposes a complex claim into a reasoning program whose steps are
delegated to sub-task functions, making the inference structure explicit
and executable~\cite{pan2023programfc}. The same idea has been specialised
to knowledge graphs, where the reasoning program is composed of predefined
executable graph functions~\cite{hao2025pgr}. Logic-LM goes further, translating
the problem into a symbolic formulation, discharging it to a deterministic
solver, and using solver error messages to drive
self-refinement: reporting gains of 39.2 points over standard prompting
and 18.4 over chain-of-thought~\cite{pan2023logiclm}. Related work
autoformalises natural language into first-order logic for fallacy
detection~\cite{lalwani2024fol} and improves the reliability of that
translation~\cite{draftprune2026}. Benchmark work has begun to separate
the factuality of a claim's content from the factuality of its logical
structure, on the argument that current pipelines ignore logical
dependencies between facts~\cite{xie2026lorefact}. It supplies a dataset
and an evaluation framework rather than an independently checkable
artefact, and we code it accordingly.

This is also where the sharpest methodological risk sits. A solver's
guarantee attaches to the formula it was given, not to the sentence a human
wrote. Where autoformalisation silently mistranslates, the pipeline emits a
\emph{confidently wrong certificate}, worse than none because it survives
scrutiny a bare prediction would not. Recent work quantifies this failure mode in legal
reasoning~\cite{knowyourlimits2026} and exposes logical flaws in
LLM-generated derivations via automated theorem
proving~\cite{beyondcorrectness2025}. We identified no equivalent study for fact-checking. A deeper objection has been raised against the whole programme, and a
survey that argued for soundness without answering it would be
incomplete. Chan et al.\ observe that a claim can be logically sound and
still misleading, because what a reader infers exceeds what the premises
support. Soundness is a semantic property, and misleadingness is a
pragmatic one~\cite{chan2026position}. Since detecting technically true but
misleading claims is part of the professional task
(Section~\ref{sec:process}), this is a real boundary on what the methods
surveyed here can deliver, and we treat it as such rather than as an
objection to be dismissed. It also bears on Section~\ref{sec:open}: a
decomposition or formalisation that is truth-preserving may still not be
implicature-preserving. Where the claim is arithmetic rather than
rhetorical the objection has less force, and the gains are correspondingly
concrete: VeriFin grounds financial claims in filed data and discharges
them to an SMT solver, accepting none of 600 incorrect test claims where
baselines accepted between 6 and 92~\cite{verifin2026}.

A parallel concern applies to claim
decomposition~\cite{min2023factscore,credence2026}, where the empirical
picture is now explicitly mixed. Decomposition trades accuracy gains
against the noise it introduces, and can burden rather than help downstream
verification~\cite{hu2025decomposition}. Our corpus contains no soundness
criterion under which decomposing a claim into atoms preserves truth.

\section{L3: Formalising the System}
\label{sec:l3}

Levels L1 and L2 ask whether the reasoning is sound. L3 asks whether the
software performing it is. Almost every technique in this section was
developed for another domain. The survey's contribution here is to identify
what transfers and, more importantly, what does not.

\subsection{Neural network verification}

Most current fact-checking pipelines contain at least one learned
component~\cite{guo2022survey,zeng2021survey}. A synthetic-media detector
decides whether an image was manipulated, a stance or entailment model
decides whether a retrieved passage supports the claim, and a language
model writes the justification. Each is a function an adversary can probe,
and the failures noted in Section~\ref{sec:intro} are failures of exactly
these components. Neural network verification is the only technique here
that answers whether \emph{any} input within a declared set can change such
a component's output, and answers it with a certificate rather than a
test-set score. It is the L3 question asked of the part of the pipeline
that emits the verdict. Where a pipeline is purely symbolic, as in the
graph-based methods of Section~\ref{sec:l1}, the question does not arise
and the L2 techniques apply instead.

Complete verification of ReLU networks became practical with SMT-based
methods~\cite{katz2017reluplex,katz2019marabou} and was scaled by
bound propagation with branch and bound, of which
$\alpha,\!\beta$-CROWN~\cite{abcrown2021} is the exemplar. Abstract
interpretation~\cite{singh2019deeppoly} trades completeness for speed. A
common specification format and an annual competition make certified
robustness a reproducible engineering claim~\cite{vnnlib,vnncomp2025}. Verification has been extended
beyond classification to object detection~\cite{objdet2024} and to
quantized networks~\cite{qvip2022}, both of which matter here. Deployed
media-forensics models are quantized, and localisation matters for
manipulation detection.

The closest template to our domain is certified robustness for network
intrusion detection~\cite{nids_sac2025}: a security classifier, an
adversarial deployment, a certified claim.

\subsection{Why $L_\infty$ is the wrong specification}

Transferring certified robustness here without changing the specification
is nonetheless a category error. Certification bounds behaviour over a
perturbation set, and the standard set is an $L_p$ ball. The adversary in an
influence operation paraphrases, recontextualises authentic material,
re-crops, re-encodes and re-captions. None of that lies in a small
$L_\infty$ ball, so a certificate over one is a genuine guarantee about a
threat model covering only a limited subset of the relevant
manipulations.

The NLP community has partly answered this. Certified robustness to
adversarial \emph{word substitutions} is an established line. Interval
bound propagation over a synonym set yields provable guarantees against
every substitution in a declared family~\cite{jia2019certified,%
huang2019verified}, and structure-free variants scale
further~\cite{ye2020safer}. So the perturbation set need not be a pixel
ball, and a survey that claimed otherwise would be wrong.

What remains uncertified is the move that actually characterises influence
operations: \emph{recontextualisation}. An authentic photograph from
another event, a genuine quotation with its qualifying sentence removed, a
real statistic attached to the wrong population. None of these perturbs the artefact at all. They alter the relation
between the artefact, the claim it is offered for, and the context it is
placed in. An \emph{artefact-local} perturbation specification cannot
express that. What information-integrity robustness needs is a relational
specification over claim, evidence and context, which is not what current
certification tooling takes as input. This is a specification
problem before it is a verification problem, and it is the reason
$L_p$ certificates and substitution certificates alike leave an important class of
integrity threats untouched. Certification of data poisoning via mixed-integer
programming~\cite{poisoning2026} is a useful adjacent case, since detectors
are retrained on adversary-influenced data.

\subsection{Distribution-free statistical guarantees}

A checker that cannot be certified can still be made to say when it should
not answer, and to bound how often it is wrong when it does.
Conformal prediction supplies distribution-free coverage under
exchangeability~\cite{vovk2005alrw}, extended by conformal risk control to
user-specified losses~\cite{angelopoulos2023crc}. Conformal factuality selects the most specific claim in a back-off chain
that retains a correctness guarantee~\cite{mohri2024conformal}, and
conformal abstention turns uncertainty into a calibrated
refusal~\cite{conformalabstention2025,onlineconformal2025}.

The framing we propose, and which the literature has not articulated, is
that these are two points on a single trade-off. The two make \emph{different} assurance trades rather than
stronger and weaker versions of one. Formal verification gives a universal
guarantee over an explicitly specified model or perturbation set. Conformal
methods give probabilistic coverage or risk control under distributional
assumptions such as exchangeability. Neither orders the other. A deployed
system may use them complementarily: certification where a specification
can be written, conformal risk control elsewhere, and an
explicit statement of which regime covers which component.

\subsection{Runtime verification}

Where whole-system static verification is currently impractical, as it is
for pipelines containing a large language model, runtime verification
supplies complementary assurance rather than a replacement. TemporalGuard treats an LLM conversation as an
execution trace, grounds messages into atomic propositions, and checks
past-time LTL safety policies over the resulting
trace~\cite{temporalguard2025}. Related work surveys runtime verification for LLM-based
autonomous systems~\cite{watchdogs2025} and develops causal past logics for
distributed agent workflows~\cite{causalpast2026}. The classical monitoring
toolchain~\cite{basin2015monpoly,tessla2018} applies directly, and monitors
can be synthesised from natural-language specifications.

This is the most immediately deployable material in the survey, though not
all of it lives in one logic. ``No silent failure in crisis mode'' and ``a
drift alarm fires within $k$ steps'' are past-time LTL properties. A
bounded false-positive \emph{rate} over a rolling window is not: Boolean
ptLTL has no numeric aggregation, and such properties need a finite
encoding, or more naturally a metric first-order logic with
aggregation~\cite{basin2015monpoly} or a stream language with numeric
computation~\cite{tessla2018}.

\subsection{Explanation soundness}

Practitioners require a fact-check to be replicable from its stated
sources~\cite{ifcn_code}, and want explanations that identify the evidence
used rather than post-hoc narratives~\cite{showmethework2025}. A proposed
evaluation framework stops short of entailment
checking~\cite{kotonya2024evaluating}. The corresponding formal property is
entailment of the published justification by the evidence actually
consulted, $E_{\text{used}} \models J$, which is decidable for the
restricted fragments of Section~\ref{sec:l1} and is, to our knowledge,
never checked. ProoFVer~\cite{krishna2022proofver} and
ExFaKT~\cite{gadelrab2019exfakt} achieve the property by construction for
their respective fragments. We identified no general-purpose mechanism for checking it across
heterogeneous fact-checking pipelines.

\subsection{Provenance and interchange}

Some integrity questions need not be inferred at all. Where a piece of media
came from can be recorded when it is created, and a published verdict can be
emitted in a form other systems can read. Content provenance is a cryptographic rather than statistical route to
integrity, and C2PA is its de facto standard~\cite{c2pa_spec}. Formal
analysis has already proved its worth: generators and validators agree on a
claim's assertions but \emph{not} on its trusted timestamp, letting an
adversary cast doubt on genuine provenance, and revocation handling is
weaker still~\cite{c2pa_security,c2pa_short2026}.
Complementary work combines metadata, watermarking and
cryptography~\cite{provenance_broadcast2024,deepfake_solutions2024}.

At the other end of the pipeline sits schema.org's
\texttt{ClaimReview}~\cite{claimreview}. Its \texttt{itemReviewed} may be a
\texttt{Claim} object, so the vocabulary does admit claims as first-class entities, but no canonical claim identifier is required, rating scales are
not semantically harmonised across organisations, and relations between
claims have thin formal semantics. In practice the relation ``these two
organisations checked the same claim and disagreed'' is expressible only by
convention, not by the standard.

Structured threat-intelligence formats have moved further:
DISARM supplies an ATT\&CK-style TTP taxonomy for influence
operations~\cite{disarm}, mapped to STIX~2.1~\cite{stix21} and exchanged
through open platforms~\cite{sanchez2025interop,sanchez2025disinfox}, with
narrative-level indicators proposed as the durable
signal~\cite{cotroneo2025fakecti}. Ontological grounding of STIX has been
demonstrated in adjacent domains~\cite{gridstix2025}.

\section{L4: Formalising the Ecosystem}
\label{sec:l4}

Levels L1 to L3 concern one claim and the system that judges it. L4 changes
the object to the environment the claim moves through, where the questions
are what a population of agents believes, how a claim spreads among them,
and whether that spread can be steered or forced.

\subsection{Epistemic and doxastic logics}

The first of those questions is about belief, and belief changes when an
agent hears something. Dynamic epistemic logic models belief change under announcement and
observation~\cite{vanbenthem2007dynamic,baltag2008qualitative}, specialised
to social networks and cascades~\cite{christoff2015logic,termmodal2019}. The bridge to diffusion proper is a dynamic-epistemic treatment of
threshold models, sound and complete for the threshold
dynamics~\cite{baltag2019diffusion}. The line continues with a formal
model of polarisation under confirmation bias~\cite{polarization2021},
quantitative variants~\cite{quantitative_del2026}, and a proof theory and
relational semantics for agents who propagate information on the basis of
distrust rather than evidence~\cite{prandi2022paranoid}. A recent synthesis
surveys the wider programme~\cite{belardinelli2026sociallogic}. Across our
corpus this work is philosophically careful but empirically disconnected,
and almost never calibrated against measured diffusion.

\subsection{Probabilistic model checking of diffusion}

The question a platform actually faces is quantitative. How likely is a
cascade of a given size, and does an intervention change that probability?
Diffusion is a stochastic process on a graph and so a target for
probabilistic model checking~\cite{kwiatkowska2011prism,dehnert2017storm},
and the formal literature here is small but not singular. Dennis et al.\
model-check a Markov-chain formalisation of diffusion in
PRISM~\cite{diffusion2022markov}, reporting candidly that even simple
models proved intractable at interesting scale and falling back on Monte
Carlo. Aldini specifies fake-news spreading in a process algebra and
analyses it in a probabilistic verification framework built on modal logic
and model checking~\cite{aldini2022algebraic}. Fionda derives temporal
formulas automatically from observed true and fake diffusion graphs and
uses their satisfaction to discriminate the two~\cite{fionda2025ndtl}.
Epidemiological formulations~\cite{misinfo_epidemic2024,misinfo_epidemic2025}
supply models but are analysed by simulation rather than verification.

Diffusion can therefore be formally modelled, and its patterns formally
classified. The step we did not find taken is to state \emph{intervention
guarantees} as properties. A media-literacy claim of the form ``if
prebunking reaches $x\%$ of a community before exposure, the probability of
a cascade exceeding size $N$ is below $p$'' is a PCTL formula over a
calibrated model, and we found no such application. It would turn literacy
policy advice from an assertion into a checkable statement, and is among
the most immediately testable gaps our corpus exhibits.

\subsection{Strategic and multi-agent verification}

Diffusion models assume agents that spread rather than agents that scheme.
The remaining L4 question is whether coordinated accounts can force an
outcome. MCMAS verifies temporal, epistemic and strategic properties of multi-agent
systems symbolically, with limited strategy
synthesis~\cite{lomuscio2017mcmas}. Detection has already been posed as a
model-checking problem. A temporal network logic specifies posting and
following behaviour, expresses bot behaviour types as formulas, and admits
model checking in polynomial time, \textsc{pspace} for its hybrid
extension~\cite{pedersen2023bots}. The \emph{strategic} question is the one
we found unasked. Whether a coalition of $k$ inauthentic accounts can force
an outcome regardless of the platform's moderation policy is an ATL
model-checking problem we identified nowhere in our corpus. Empirically, coordinated inauthentic behaviour is detected through
synchrony and latent coordination
networks~\cite{cib_likes2023,coordinated2025}, and temporal graph patterns
have been given a timed-automata semantics~\cite{tgpattern2023} that offers
a formal query language for exactly these structures.

\section{L5: Compliance as Verification}
\label{sec:l5}

The DSA obliges very large platforms to assess systemic risks and submit
to independent annual audits~\cite{dsa2022}. The AI Act requires conformity
assessment for designated high-risk systems~\cite{aiact2024}. Both impose
obligations that function as specifications without being written in a
specification language, and audit practice is correspondingly
heterogeneous. Analyses of audit methodology find
significant inconsistency and limited technical depth when AI systems are
evaluated~\cite{dsa_blindspot2026,ojewale2024audits}, and recent work
begins the translation from legal requirement to technical
verification~\cite{airact_verification2025}.

Formal methods produce one class of artefact that regulatory assurance
increasingly needs: machine-checkable evidence that a precisely stated
technical obligation has been discharged. They cannot discharge the
organisational, procedural and human-governance evidence that also
counts. It applies inside the
profession as well as outside it. As noted in Section~\ref{sec:process}, four of
the IFCN's five commitments constrain procedure, and a procedural
constraint is a monitorable property.

\section{Cross-Cutting Analysis}
\label{sec:analysis}

\paragraph*{What the systematization changes} Three intuitions that are
reasonable in isolation do not survive the cross-literature comparison.
First, the shortage is not of formal machinery but of its adoption under
information-integrity specifications. A substantial pool of relevant
techniques is mature and was built elsewhere (F1), while the stages that
would use it lack formalised operations or correctness criteria (F2). Second, a stronger guarantee is
not automatically a more relevant one, because a certificate can be sound
and still range over the wrong perturbation set, answering a question
nobody asked (Section~\ref{sec:l3}). Third, the field's accuracy benchmarks
cannot separate a system that derives a verdict from one that predicts the
same label, so the property that distinguishes the formal approach is the
one not being scored (F3). The rest of this section establishes each from
the coded corpus.

Table~\ref{tab:counts} reports the coded corpus by level and role.
Four findings follow, each stated so that the evidence for it is a
number in that table rather than an impression from the prose.

\begin{table}[t]\centering
\caption{The coded corpus: \CorpusN{} works, each assigned to the level of
its primary object and counted once, so the levels partition the corpus and
the totals are exact. Every figure here is generated from the
coding file by script, not maintained by hand. The bibliography carries \BibN{} entries; the difference is works
cited only outside Section~\ref{sec:l1}--Section~\ref{sec:l5}, which are not coded.}
\label{tab:counts}
\footnotesize
\setlength{\tabcolsep}{4pt}
\begin{tabular}{@{}llrrrr@{}}
\toprule
& \textbf{Level} & \textbf{Direct} & \textbf{Transfer} & \textbf{Context}
& \textbf{All} \\
\midrule
L1 & The object     & \NLOneDirect   & \NLOneTransfer            & \NLOneContext   & \NLOne \\
L2 & The reasoning  & \NLTwoDirect   & \NLTwoTransfer            & \NLTwoContext   & \NLTwo \\
L3 & The system     & \NLThreeDirect & \textbf{\NLThreeTransfer} & \NLThreeContext & \NLThree \\
L4 & The ecosystem  & \NLFourDirect  & \NLFourTransfer           & \NLFourContext  & \NLFour \\
L5 & Governance     & \NLFiveDirect  & \NLFiveTransfer           & \NLFiveContext  & \NLFive \\
\midrule
& \textbf{Total}   & \NTotDirect & \textbf{\NTotTransfer} & \NTotContext & \CorpusN \\
\bottomrule
\end{tabular}
\end{table}

\begin{table}[t]\centering
\caption{The assurance landscape: what is formalised (rows)
against the artefact the method produces (columns). Each cell
reads \textbf{direct}/transfer over the \CorpusN{} coded works, and
\textnormal{---} marks a pairing absent from the corpus. Shading
tracks the transfer count, so a dark cell with a small bold number
is the adoption gap in one glance. Context items yield no artefact
and do not appear, and every other work appears exactly once, so the
entries sum to the totals of Table~\ref{tab:counts}.}
\label{tab:matrix}
\footnotesize
\setlength{\tabcolsep}{3pt}
\begin{tabular}{@{}ll ccccc@{}}
\toprule
& & \textbf{Proof} & \textbf{Witness} & \textbf{Cert.} & \textbf{Quant.} & \textbf{Runtime} \\
\midrule
L1 & The object      & \cellcolor{dB}\textbf{8}/2 & \cellcolor{dA}\textbf{5}/0 & \cellcolor{dA}{\color{black!35}---} & \cellcolor{dA}{\color{black!35}---} & \cellcolor{dA}{\color{black!35}---} \\
L2 & The reasoning   & \cellcolor{dE}\textbf{9}/14 & \cellcolor{dC}\textbf{0}/3 & \cellcolor{dA}{\color{black!35}---} & \cellcolor{dC}\textbf{0}/4 & \cellcolor{dA}{\color{black!35}---} \\
L3 & The system      & \cellcolor{dB}\textbf{0}/1 & \cellcolor{dA}\textbf{2}/0 & \cellcolor{dE}\textbf{1}/11 & \cellcolor{dC}\textbf{1}/4 & \cellcolor{dC}\textbf{0}/5 \\
L4 & The ecosystem   & \cellcolor{dD}\textbf{0}/7 & \cellcolor{dB}\textbf{1}/2 & \cellcolor{dA}{\color{black!35}---} & \cellcolor{dC}\textbf{2}/3 & \cellcolor{dA}\textbf{1}/0 \\
L5 & Governance      & \cellcolor{dA}{\color{black!35}---} & \cellcolor{dA}{\color{black!35}---} & \cellcolor{dA}{\color{black!35}---} & \cellcolor{dA}{\color{black!35}---} & \cellcolor{dB}\textbf{0}/1 \\
\bottomrule
\end{tabular}
\end{table}

\begin{table}[t]\centering
\caption{When the corpus was written. Three undated living
standards (the IFCN code, the C2PA specification,
\texttt{ClaimReview}) carry no publication year; they are omitted
from the period columns and retained in the totals. The
concentration in 2024--26 is sharpest at L3, which is the
quantitative basis for the claim that the field's centre of
gravity has moved recently.}
\label{tab:era}
\footnotesize
\setlength{\tabcolsep}{4pt}
\begin{tabular}{@{}ll rrrr r@{}}
\toprule
& & $\le$\textbf{2015} & \textbf{2016--20} & \textbf{2021--23} & \textbf{2024--26} & \textbf{All} \\
\midrule
L1 & The object      & 3 & 5 & 4 & 8 & 20 \\
L2 & The reasoning   & 9 & 7 & 9 & 12 & 37 \\
L3 & The system      & 1 & 8 & 4 & 22 & 38 \\
L4 & The ecosystem   & 4 & 4 & 6 & 7 & 21 \\
L5 & Governance      & 0 & 0 & 1 & 4 & 5 \\
\midrule
& \textbf{Total} & 17 & 24 & 24 & 53 & \CorpusN \\
\bottomrule
\end{tabular}
\end{table}

\textbf{F1: the adoption gap.} Counted as raw activity the distribution
peaks at L3, with \NLThree{} works, which reads as a healthy literature.
The role coding says otherwise. Of those \NLThree{} entries,
\NLThreeDirect{} are formal or certifiable methods applied \emph{directly}
to information integrity, \NLThreeTransfer{} are transfer candidates built
for other domains, and \NLThreeContext{} are context items. The distance
between \NLThreeTransfer{} available and \NLThreeDirect{} instantiated is
the finding, and it is not local to L3. Across the whole corpus
\NTotTransfer{} works are transfer candidates against \NTotDirect{} direct
applications. The machinery exists and this field has largely not taken it
up. Read as native activity rather than as available tooling, L1 and L2 are
the mature levels, with \NLOneDirect{} and \NLTwoDirect{} direct works
respectively, while L3, L4 and L5 hold \NLThreeDirect{}, \NLFourDirect{}
and \NLFiveDirect{}. That sparsity is not a consequence of any logical
dependency on the levels below. It is simply where the field has not gone.
This answers RQ2, and Table~\ref{tab:matrix} answers RQ1 in the same
stroke by crossing the levels with the artefact each method yields. Read
it for the dark cells with small bold numbers. Certificate-producing
methods at L3 stand \NLThreeCertDirect{} instantiated against
\NLThreeCertTransfer{} transfer candidates, and the
\NLFourProofTransfer{} proof-producing works at L4 are transfer without
exception. Read it also for the dashes, which are the genuinely unoccupied
combinations: no certificate-producing work at L1, L2, L4 or L5, and
nothing but a single transfer candidate at L5 at all. The empty cells are
a research agenda stated as a picture.

\textbf{F2: the specification gap.} The \emph{specification} gaps of
Section~\ref{sec:mismatch} concern stages that exist in the pipeline: S4 and S8
lack a formalised operation, S3 and S7 lack a correctness criterion. A
different gap is one of \emph{coverage}: \emph{Intervene} (S9) attracts
essentially no directly instantiated work at any level, even though L4
contains formal models of intervention effects. The apparatus is built to
justify interventions that nothing in it formally treats.

\textbf{F3: the evaluation gap.} The standard evaluation
resources (FEVER, FEVEROUS, HOVER, SciFact, TabFact, AVeriTeC and
QuanTemp~\cite{thorne2018fever,aly2021feverous,jiang2020hover,%
wadden2020scifact,chen2020tabfact,schlichtkrull2023averitec,quantemp2024,averimatec2026}) %
are inventoried exhaustively elsewhere~\cite{guo2022survey}, so we do not
reproduce them. The relevant observation is what they all lack. Not one
carries ground-truth derivations, specifications, perturbation sets or
provenance chains; TabFact pairs tables with statements and entailment labels. The logical
forms in that line are \emph{derived} by systems such as
LogicalFactChecker rather than supplied as benchmark ground
truth~\cite{zhong2020logicalfactchecker}. A system that emits a proof and a system that emits a label are
therefore scored identically.

The point is sharper in the field's own evaluation infrastructure.
OpenFactCheck ships a fact-checker leaderboard whose reported dimensions
are \emph{accuracy, latency and cost}~\cite{iqbal2025openfactcheck}. Those
are the right three axes for procuring a checker and the wrong three for
telling a sound checker from an unsound one. A system whose verdict is
entailed by its stated evidence and a system whose verdict merely
correlates with it occupy the same leaderboard position. This is not a minor gap; it is why L2
progress is measured in accuracy points rather than in the property that
actually distinguishes the approach.

\textbf{F4: the composition gap.} Every guarantee in the corpus attaches
to a component. A certified detector, a replayable derivation and a
runtime monitor are each available in isolation, and we identified no
framework in which they compose into a statement about the pipeline that
produced a published verdict. Table~\ref{tab:warrant} makes the
consequence concrete by carrying one claim down the stack. The artefacts
exist at L2 and partly at L3, and the two levels a reader would most want,
the scoped claim and the intervention guarantee, are the two the
literature does not supply. That is the composition gap and the
specification gap meeting on a single example.

\begin{table*}[t]\centering
\caption{One claim carried through the stack: ``Unemployment fell by 10\%
in Cyprus during 2025.'' Each row gives the question its level asks, the
artefact that would answer it, and whether the corpus contains a method
producing that artefact today. The warrant is the tuple of artefacts in
the third column, and the two rows a reader would most want are the two
the literature does not supply.}
\label{tab:warrant}
\footnotesize
\setlength{\tabcolsep}{5pt}
\renewcommand{\arraystretch}{1.15}
\begin{tabular}{@{}p{15mm}p{47mm}p{55mm}p{45mm}@{}}
\toprule
& \textbf{What this level asks} & \textbf{Artefact that answers it}
& \textbf{Available today?} \\
\midrule
L1 object
& Which proposition is asserted? ``Fell by 10\%'' of what population, over
  which interval, on which measure, against which data revision?
& A typed claim with scope fixed: Cyprus registered labour force,
  2025-01-01 to 2025-12-31, seasonally adjusted rate, Eurostat vintage.
& \textbf{No.} This is stage S4, which we found has no formalised
  operation at all (Section~\ref{sec:mismatch}); the nearest work fixes
  quantification for graph claims~\cite{hao2026folxkg}. \\
L2 reasoning
& Does the evidence entail the claim under that scope?
& A derivation $E \vdash C$: two series values, one relative-change
  computation, replayable by an independent checker.
& \textbf{Yes.} Arithmetic claims of this shape are exactly where solver
  grounding works~\cite{verifin2026}. \\
L3 system
& Is the checker that produced the verdict itself trustworthy?
& $E_{\text{used}} \models J$ for the published justification; a
  robustness certificate over paraphrases of the claim; a provenance
  manifest for any chart image.
& \textbf{Partly.} Provenance is standardised~\cite{c2pa_spec} and
  justification entailment is decidable here, but we identified no
  general mechanism that checks it (Section~\ref{sec:l3}). \\
L4 ecosystem
& If the correction is published, does it change what people believe?
& $\Pr[\text{cascade} > N] < p$ given prebunking reach $x\%$: a PCTL
  property over a calibrated diffusion model.
& \textbf{No.} Diffusion is formally modelled and
  classified~\cite{aldini2022algebraic,fionda2025ndtl}; intervention
  guarantees are not stated as properties (Section~\ref{sec:l4}). \\
L5 governance
& Which audit obligation does this evidence discharge?
& The bundle above, addressed to a named DSA risk-assessment duty.
& \textbf{No.} The obligation is not written in a specification language
  (Section~\ref{sec:l5}). \\
\bottomrule
\end{tabular}
\end{table*}

One deployed artefact sits outside the corpus but is worth naming, because
it is what the warrant notion is set against. A common deployed answer to
``should I believe this source'' is a reputation mechanism, a
source-credibility rating or platform trust signal. Measured against
Table~\ref{tab:guarantees} such a score yields none of the five artefacts.
There is no derivation to replay, no witness localising which past
judgement drove the rating, no declared perturbation set over which it is
stable, no calibration statement licensing it as a probability, and no
property of which it is a verdict on a run. It is a bare verdict moved up
one level, from the claim to the source, and the DSA gives such judgements
procedural force through trusted-flagger status~\cite{dsa2022}. The formal
counterpart already appears above, since truth discovery jointly infers
source reliability and claim veracity
(Section~\ref{sec:l2})~\cite{dong2009truthdiscovery,li2016truthsurvey}, but
the deployed mechanisms are not connected to it.

\section{Open Problems}
\label{sec:open}

\begin{table}[t]\centering
\caption{Provenance of the research agenda. Each open problem of
Section~\ref{sec:open} is a deduction from a specific observation in the coded
corpus, not a free-standing suggestion. The middle column names the gap
type: \textsc{spec}ification, \textsc{cov}erage, \textsc{ad}option,
\textsc{ev}aluation or \textsc{comp}osition.}
\label{tab:agenda}
\footnotesize
\setlength{\tabcolsep}{4pt}
\begin{tabular}{@{}p{46mm}p{11mm}l@{}}
\toprule
\textbf{Observation in the corpus} & \textbf{Gap} & \textbf{Problem} \\
\midrule
S4 has no formalised operation (Section~\ref{sec:mismatch})            & \textsc{spec} & 1 \\
Autoformalisation faithfulness unstudied here (Section~\ref{sec:l2})   & \textsc{spec} & 2 \\
L3 certificates declare unsuitable perturbation sets             & \textsc{spec} & 3 \\
No completeness property for evidence sets                       & \textsc{cov}  & 4 \\
Decomposition has no preservation criterion                      & \textsc{comp} & 5 \\
Certificates attach to components, not pipelines                 & \textsc{comp} & 6 \\
Refuted and unsupported are conflated                            & \textsc{spec} & 7 \\
L4 model checking intractable at real scale                      & \textsc{ad}   & 8 \\
Certificate strength untested on non-experts                     & \textsc{ev}   & 9 \\
No benchmark scores derivations (Section~\ref{sec:analysis})           & \textsc{ev}   & 10 \\
L5 holds \NLFiveDirect{} direct works (Table~\ref{tab:counts})   & \textsc{ad}   & 11 \\
\bottomrule
\end{tabular}
\end{table}

\emph{1. The specification problem.} Progress here is real but partial:
first-order logic over knowledge graphs has been extended with comparison
predicates and counting quantifiers, with claims paired to formulas and an
executable prover~\cite{hao2026folxkg}, which settles that quantification,
comparison and counting can be made explicit for graph-grounded claims.
What remains missing is the rest of the scope, namely temporal extent,
population, measure, source version and evidence context, and we
identified no generally adopted specification language for stating
``this claim is verifiable'' or ``this verdict is sound''. Everything else
here is downstream. \emph{First step:} a specification language for claim
scope (quantifier, interval, population, measure, data revision) with a
decidable satisfaction relation.

\emph{2. Autoformalisation faithfulness.} Silent mistranslation converts
a solver certificate into confident nonsense. \emph{First step:} replicate
the legal-domain faithfulness study~\cite{knowyourlimits2026} on
FEVEROUS and AVeriTeC, and report translation error separately from
end-task accuracy.

\emph{3. The wrong perturbation set.} $L_\infty$ certificates do not
speak to paraphrase or recontextualisation, and the harder case is
relational. The artefact may be untouched while the relation between claim,
evidence and context changes. \emph{First step:} define a \emph{relational}
perturbation model over (claim, evidence, context) admitting bounded
paraphrase but also context substitution, quotation truncation, population
reassignment and temporal relocation, then establish whether it admits
tractable sound over-approximation.

\emph{4. Assurance for retrieval.} Verdict soundness is vacuous when
counter-evidence is systematically missing~\cite{glockner2022missing}.
\emph{First step:} a coverage contract relative to a \emph{declared}
source universe, query policy and time horizon, of the form ``every
qualifying document in this source set up to time $t$ was considered'',
with a monitor that detects its violation at runtime. Global completeness
over the open Web is not establishable and is the wrong target.

\emph{5. Truth-preserving decomposition.} We identified no accepted
criterion in our corpus under which splitting a claim into atoms preserves
truth conditions. \emph{First step:}
state the property in natural logic and test whether existing decomposers
satisfy it.

\emph{6. Compositional assurance.} Component certificates do not compose
into a pipeline guarantee. \emph{First step:} assume-guarantee contracts at
neural/symbolic boundaries.

\emph{7. Open-world verification.} Absence of proof is routinely
conflated with proof of absence. \emph{First step:} an explicit epistemic
treatment distinguishing ``refuted'' from ``unsupported''.

\emph{8. Scaling L4.} PRISM is intractable at realistic network
sizes~\cite{diffusion2022markov}. \emph{First step:} mean-field abstraction
or statistical model checking with stated error bounds.

\emph{9. Certificates for citizens.} A guarantee no non-expert can
interpret has no democratic value. \emph{First step:} study how
practitioners and readers interpret certificate
strength~\cite{showmethework2025}.

\emph{10. No benchmark.} We identified no shared task centred on machine-checkable formal
assurance, and no harness that scores the property distinguishing it.
\emph{First step:} add a derivation-validity dimension to an existing
harness rather than building one. OpenFactCheck's
\textsc{CheckerEval}~\cite{iqbal2025openfactcheck} is the natural host,
since it already accepts pluggable verifiers.

\emph{11. Regulation as specification.} What does a conformity assessment
for an information-integrity system contain? \emph{First step:} a
machine-checkable rendering of one DSA audit obligation, end to end.

\section{Conclusion}

Automated fact-checking is optimised for producing verdicts. The warrants
that would make those verdicts checkable remain exceptional rather than a
standard output. Organising the formal-methods literature by what
it formalises shows why. Direct adoption drops sharply beyond reasoning:
L1 and L2 hold \NLOneDirect{} and \NLTwoDirect{} direct works, while L3 and
L4 hold \NLThreeDirect{} and \NLFourDirect{} apiece alongside
substantially larger pools of transferable machinery, and L5 remains almost
entirely contextual at \NLFiveDirect{} direct works. Mature tooling exists for
many of the L3 and L4 assurance problems, built for adjacent domains under
comparable adversarial conditions, while translating governance obligations
into technical verification remains nascent. We found no generally adopted
specification of what an automated fact-checking system should guarantee
beyond task-level correctness, so the central obstacle is specification
rather than absent machinery. Scalability, formalisation faithfulness and
evidence completeness are obstacles too. Regulation increasingly makes the
demand for auditable evidence operational, and the professional practice
the field automates has, in its own code of conduct, already articulated
part of the specification.

\balance
\bibliographystyle{IEEEtran}
\bibliography{refs}

\end{document}